\pdfoutput=1
\documentclass[11pt]{article}

\usepackage[]{EMNLP2023}

\usepackage{times}
\usepackage{latexsym}
\usepackage{amsmath}
\usepackage{graphicx}
\usepackage{subfig}
\usepackage{color}
\usepackage{arydshln}
\usepackage{multirow}
\usepackage[T1]{fontenc}
\usepackage[utf8]{inputenc}

\usepackage{microtype}

\usepackage{inconsolata}

\usepackage{algorithm}
\usepackage{algpseudocode}

\ifx\figurename\undefined \def\figurename{Figure}\fi
\ifx\figurenames\undefined \def\figurenames{Figures}\fi
\ifx\tablename\undefined \def\tablename{Table}\fi
\ifx\tablenames\undefined \def\tablenames{Tables}\fi
\ifx\equationname\undefined \def\equationname{Equation}\fi
\ifx\equationnames\undefined \def\equationnames{Equations}\fi

\title{The Limits of ChatGPT in Extracting Aspect-Category-Opinion-Sentiment Quadruples: A Comparative Analysis}

\author{Xiancai Xu\footnotemark[1] \and Jia-Dong Zhang\footnotemark[1]~~\footnotemark[2] \and Rongchang Xiao \and Lei Xiong\\
         Brands \& Consumers Research Institute, Enbrands Inc., Shenzhen, China \\ 
         \{essen, zhangjd.1, xiaorc.1, xiongl.1\}@enbrands.com}
\begin{document}
\maketitle
\renewcommand{\thefootnote}{\fnsymbol{footnote}}
\footnotetext[1]{Equal contribution.}
\footnotetext[2]{Corresponding author.}

\begin{abstract}
Recently, ChatGPT has attracted great attention from both industry and academia due to its surprising abilities in natural language understanding and generation. We are particularly curious about whether it can achieve promising performance on one of the most complex tasks in aspect-based sentiment analysis, i.e., extracting aspect-category-opinion-sentiment quadruples from texts. To this end, in this paper we develop a specialized prompt template that enables ChatGPT to effectively tackle this complex quadruple extraction task. Further, we propose a selection method on few-shot examples to fully exploit the in-context learning ability of ChatGPT and uplift its effectiveness on this complex task. Finally, we provide a comparative evaluation on ChatGPT against existing state-of-the-art quadruple extraction models based on four public datasets and highlight some important findings regarding the capability boundaries of ChatGPT in the quadruple extraction. 
\end{abstract}
\renewcommand{\thefootnote}{\arabic{footnote}}
\section{Introduction}
Since its release by OpenAI in November 2022, ChatGPT\footnote{We do not include GPT-4 in this study due to its expensive API, but ChatGPT still offers valuable insights.}, as a representative of large language models (LLMs), has persistently received ``overwhelming'' attention from both industry and academia. ChatGPT uses reinforcement learning from human feedback (RLHF)~\cite{Ouyang_Wu_Jiang_2022} to align itself with human preference and generates fluent responses for various user queries. This universal paradigm enables ChatGPT to achieve remarkable performance on content generation, e.g., writing essay and code. ChatGPT has been widely applied in a variety of natural language processing (NLP) tasks with zero-shot learning~\cite{Gao_Wang_Hou_2023,Zhong_Ding_Liu_2023,Wang_Xie_Ding_2023,Wei_Cui_Cheng_2023,Yuan_Xie_Ananiadou_2023}. These works have also reported that the performance can be significantly improved based on few-shot prompting by providing a few examples for ChatGPT to conduct in-context learning~\cite{Brown_Mann_Ryder_2020}.

\begin{figure}[!t]
	\centering
	\subfloat[A product review]{\label{subfig:acos-text}%
		\begin{tabular}{|@{~}p{\columnwidth}@{~}|}
			\hline
			\textbf{Original text:} \textit{``Looks {\color{blue}nice} and the {\color{red}surface} is {\color{blue}smooth}, but certain {\color{red}apps} take seconds to respond''}\\
			\hline
		\end{tabular}
	}

	\subfloat[Quadruples]{\label{subfig:acos-result}%
		\begin{tabular}{cccc}
		\hline
		\textbf{\color{red}Aspect}  & \textbf{\color{magenta}Category}  &  \textbf{\color{blue}Opinion} & \textbf{\color{cyan}Sentiment} \\
		\hline
		{\color{red}surface} & {\color{magenta}Design}    &  {\color{blue}smooth}  & {\color{cyan}Positive} \\
		\hline
		{\color{red}null}    & {\color{magenta}Design}    &  {\color{blue}nice}    & {\color{cyan}Positive} \\
		\hline
		{\color{red}apps}    & {\color{magenta}Software}  &  {\color{blue}null}    & {\color{cyan}Negative} \\
		\hline
	\end{tabular}
	}
	
	\caption{An example of the aspect-category-opinion-sentiment quadruple extraction}
	\label{fig:acos-example}
\end{figure}

The promising performance of ChatGPT has sparked our curiosity: how does ChatGPT perform on one of the most complicated tasks in aspect-based sentiment analysis, i.e., extracting aspect-category-opinion-sentiment quadruples from texts~\cite{Cai_Xia_Yu_2021}? For example, \figurename~\ref{subfig:acos-text} shows a product review \textit{``Looks nice and the surface is smooth, but certain apps take seconds to respond''}~\cite{Cai_Xia_Yu_2021}, in which ``surface'' is an aspect term and classified into the ``Design'' category, ``smooth'' is the opinion term toward this aspect with the ``Positive'' sentiment. The four elements constitute an quadruple ``surface-Design-smooth-Positive'', as shown in \figurename~\ref{subfig:acos-result}. Obviously, there are two more quadruples: ``null-Design-nice-Positive'' and ``apps-Software-null\footnote{We consider ``null'' as the opinion term rather than ``takes seconds to respond''. The reason is that ``takes seconds to respond'' does not explicitly express a sentiment polarity and we must infer the sentiment polarity from the whole review.}-Negative'', where ``null'' stands for an implicit aspect or opinion term that does not appear in the given text. Note that neither category nor sentiment can be ``null''.

To address the curiosity, this paper aims to investigate the effectiveness of ChatGPT in extracting aspect-category-opinion-sentiment quadruples with few-shot learning. First, we design a dedicated prompt template for this quadruple extraction task with five important parts including \textit{instruction}, \textit{context}, \textit{output format}, \textit{input data} and \textit{examples}. Second, we propose a method to choose a few appropriate examples in the few-shot learning; our method applies the k-nearest neighbors (KNN) algorithm to discover few-shot examples from training data for a given test example. Finally, we conduct extensive experiments on four public datasets~\cite{Cai_Xia_Yu_2021,Zhang_Deng_Li_2021} to evaluate the effectiveness of ChatGPT with our prompt template and compare it with the state-of-the-art quadruple extraction models to gain a comprehensive understanding of the capability boundaries of ChatGPT in this task. 

The contributions of this paper can be summarized as follows:
\begin{itemize}
	\item We conduct a study on prompt engineering and develop a specialized prompt template tailored to the complex task of aspect-category-opinion-sentiment quadruple extraction. Our prompt template empowers ChatGPT to comprehend and be competent to this quadruple extraction task.
	
	\item We propose a method to select few-shot examples for the in-context learning of ChatGPT. This method enables ChatGPT to accomplish competitive effectiveness against existing quadruple extraction models.
	
	\item We highlight important findings on the capability boundaries of ChatGPT in this task from our extensive experiments. These findings provide insights into the strengths and weaknesses of ChatGPT in the quadruple extraction. These insights can be used to guide future research and inform the design of more effective algorithms for extracting aspect-category-opinion-sentiment quadruples. 

\end{itemize}

The rest of this paper is organized as follows. We present some related work in Section~\ref{sec:rw}. Section~\ref{sec:pd} investigates in-context learning with a prompt template and few-shot examples, followed by the experimental evaluation in Section~\ref{sec:exp}. Finally, we conclude this paper in Section~\ref{sec:con}.

\section{Related Work}\label{sec:rw}
This section reviews aspect-based sentiment analysis (ABSA) and prompt engineering.

\textbf{Aspect-based sentiment analysis.} There are mainly six types of ABSA tasks. (1)~The aspect-opinion pair extraction is one of the most common ABSA tasks that simultaneously mines aspect terms, opinion terms, and their correspondence by applying a sequence labeling model with a BIO-based schema~\cite{Chen_Liu_Wang_2020} or a span-based scheme~\cite{Zhao_Huang_Zhang_2020,Gao_Wang_Liu_2021,Chakraborty_Kulkarni_Li_2022}. (2)~The aspect-sentiment pair extraction produces a list of aspect terms and their corresponding sentiments by employing two sequence labeling models to respectively predict aspects and sentiments~\cite{He_Lee_Ng_2019,Chen_Qian_2020} or integrating the two sequence labeling models into a unified model~\cite{Li_Bing_Li_2019}. (3)~The category-sentiment pair extraction aims to classify aspect categories instead of extracting aspect terms and predict their corresponding sentiments based on an attentive LSTM~\cite{Ma_Peng_Cambria_2018}, constrained attention networks~\cite{Hu_Zhao_Zhang_2019}, hierarchical graph convolutional networks~\cite{Cai_Tu_Zhou_2020} or the BERT model~\cite{Dai_Peng_Chen_2020}. (4)~The aspect-opinion-sentiment triplet extraction combines aspect-opinion pair extraction with aspect-sentiment pair extraction via adopting a pipeline solution~\cite{Peng_Xu_Bing_2020,Mao_Shen_Yu_2021,Xu_Chia_Bing_2021,Chen_Chen_Sun_2022}, designing a new tagging scheme for extracting aspects, opinions and sentiments simultaneously in one step~\cite{Wu_Ying_Zhao_2020,Xu_Li_Lu_2020}, or utilizing a pre-trained sequence-to-sequence model~\cite{Yan_Dai_Ji_2021}. (5)~The aspect-category-sentiment triplet extraction integrates aspect-sentiment pair extraction with category-sentiment pair extraction by encoding every category-sentiment pair into the BERT model along with the origin sentence for extracting the aspect-category-sentiment triplets~\cite{Wan_Yang_Du_2020} or developing a unified generative framework based on a pre-trained sequence-to-sequence model~\cite{Zhang_Li_Deng_2021}. (6)~The aspect-category-opinion-sentiment quadruple extraction is one of the most complex ABSA tasks. Most studies apply a sequence-to-sequence model to generate a list of quadruples~\cite{Zhang_Deng_Li_2021,Bao_Zhongqing_Jiang_2022,Mao_Shen_Yang_2022}. Further, recent studies pay attention on extracting implicit aspects~\cite{Cai_Tu_Zhou_2020,Wan_Yang_Du_2020,Zhang_Li_Deng_2021,Zhang_Deng_Li_2021,Mao_Shen_Yang_2022}, or  implicit opinions~\cite{Setiowati_Djunaidy_Siahaan_2022}. More comprehensively, the study~\cite{Cai_Xia_Yu_2021}  deals with implicit aspects and opinions simultaneously.

\textbf{Prompt engineering.} Prompt engineering is the process of designing high quality prompts of a task and offers a natural interface for users to interact with LLMs such as encoder-only BERT~\cite{Devlin_Chang_Lee_2018}, decoder-only GPT-3~\cite{Brown_Mann_Ryder_2020}, and encoder-decoder T5~\cite{Raffel_Shazeer_Roberts_2020}. As a typical decoder-only LLM, the recently released ChatGPT has attracted great attention, due to its impressive ability to generate fluent responses for a variety of NLP tasks with a proper prompt, including translation~\cite{Gao_Wang_Hou_2023}, question answering~\cite{Zhong_Ding_Liu_2023}, sentiment analysis~\cite{Wang_Xie_Ding_2023}, and information extraction~\cite{Wei_Cui_Cheng_2023,Yuan_Xie_Ananiadou_2023}. ChatGPT also shows an ability known as in-context learning~\citep{Brown_Mann_Ryder_2020} by providing a few examples along with the prompt for improving its performance. Meanwhile, by fine-tuning the LLM T5 with instructional prompts, some works~\cite{Gao_Fang_Liu_2022,Wang_Xia_Yu_2022,Varia_Wang_Halder_2022} develop a unified generative framework  for a variety of ABSA tasks including quadruple extraction and achieve remarkable effectiveness. In this work, we concentrate on exploring the ability of ChatGPT in quadruple extraction, which is important but has been given little attention.

\section{Methodology}\label{sec:pd}
This section first introduces a dedicated prompt template for the aspect-category-opinion-sentiment quadruple extraction in Section~\ref{subsec:pt} and then presents a method to discover few-shot examples from training data that are filled in the prompt template for in-context learning in Section~\ref{subsec:fse}.

\subsection{Prompt Template for Quadruple Extraction}\label{subsec:pt}
It is undeniable that ChatGPT has impressive capabilities in a wide range of zero-shot and few-shot tasks; however, its task-specific performance is heavily reliant on the quality of prompts used to guide it~\cite{Zhou_Muresanu_Han_2023}. To this end, a dedicated prompt template is designed to complete this quadruple extraction that is one of the most complicated tasks in aspect-based sentiment analysis. First, we determine the important parts. In general, a prompt includes an instruction and input data. The quadruple extraction is a new and complex task, and contains strong background and constraints. Therefore, context, output format and few-shot examples are required. Then, we follow the flow of natural language to order these five important parts: \textit{instruction}, \textit{context}, \textit{output format}, \textit{input data} and \textit{examples}, as depicted in \figurename~\ref{fig:prompt}.

\begin{figure}[!t]
	\centering
	\subfloat[The prompt template]{\label{subfig:prompt-template}%
	\begin{tabular}{|p{\columnwidth}|}
		\hline
		\textbf{Instruction:} \textit{extract aspect-category-opinion-sentiment quadruples from input data}\\
		\textbf{Context:} \textit{an aspect or opinion must be a term existing in input data or null if non-existing; the category is one in the predefined list <placeholder>; the sentiment is positive, negative or neutral; do not ask me for more information, I am unable to provide it, and just try your best to finish the task. You can learn from the following examples.}\\
		\textbf{Output format:} \textit{(aspect, category, opinion, sentiment)}\\
		\hdashline
		\textbf{Input:} \textit{Looks nice and the surface is smooth, but certain apps take seconds to respond.}\\
		\textbf{Output:} \textit{[(surface, Design, smooth, Positive), (null, Design, nice, Positive), (apps, Software, null, Negative)]}\\
		\textbf{Input:} ...... \ \ \ \ \ {\color{red}<this block for examples>}\\
		\textbf{Output:} ......\\
		\hdashline
		\textbf{Input:} \textit{<a record from test data>}\\
		\textbf{Output:}\\
		\hline
	\end{tabular}
	}

	\subfloat[A prompt example generated from the template]{\label{subfig:prompt-example}%
	\begin{tabular}{|p{\columnwidth}|}
		\hline
		Instruction: extract aspect-category-opinion-sentiment quadruples from input data\\
		Context: an aspect or opinion must be a term existing in input data or null if non-existing;\\ 
		the category is one in the predefined list: [`restaurant general', `service general', `food quality', `food style\_options', `drinks style\_options', `drinks prices', `restaurant prices', `ambience general', `restaurant miscellaneous', `food prices', `location general', `drinks quality'];\\
		the sentiment is positive, negative or neutral;\\
		do not ask me for more information, I am unable to provide it, and just try your best to finish the task.\\
		You can learn from the following examples.\\
		Output format: (aspect, category, opinion, sentiment)\\
		Input: it was really good pizza .\\
		Output: [(pizza, food quality, good, positive)]\\
		Input: the fish was really , really fresh .\\
		Output: [(fish, food quality, fresh, positive)]\\
		Input: great sushi experience .\\
		Output: [(sushi, food quality, great, positive)]\\
		Input: serves really good sushi .\\
		Output:\\
		\hline	
	\end{tabular}
	}
	
	\caption{The prompt design for extracting aspect-category-opinion-sentiment quadruples from input data}
	\label{fig:prompt}
\end{figure}

\textbf{Instruction.} The instruction is often a command that steers ChatGPT to complete a specific task. Our instruction is straightforward, i.e., \textit{``Extract aspect-category-opinion-sentiment quadruples from input data''}.  
 
\textbf{Context.} The context provides additional information, e.g., background and constraints, that helps ChatGPT to better understand a specific task and generate responses. In our case, the context is \textit{``an aspect or opinion must be a term existing in input data or null if non-existing; the category is one in the predefined list <placeholder>; the sentiment is positive, negative or neutral; do not ask me for more information, I am unable to provide it, just try your best to finish the task, and you can learn from the following examples''}, in which <placeholder> will be filled in a list of category names when the prompt template is instantiated with a concrete dataset.

\textbf{Output format.} In this extraction task, each outputted quadruple contains four elements with the format \textit{``(aspect, category, sentiment, opinion)''}. The output format is important for a complicated task to guide ChatGPT to generate useful responses. To reduce the length of the prompt, we adopt the concise structure instead of a JSON-like structure.

\textbf{Input data.} This is the exact input or question that we are seeking a response to. In our experiments, we separately consider each record in test data as the input to obtain its outputted quadruples.

\textbf{Examples.} As we know, in the zero-shot setting ChatGPT still falls short on more complex tasks, e.g., the aspect-category-opinion-sentiment quadruple extraction. Thus, we offer a few-shot learning approach by providing examples from training data that enable ChatGPT to learn in-context, resulting in improved performance. Following is an example:\\
\textit{``\textbf{Input:} Looks nice and the surface is smooth, but certain apps take seconds to respond.\\
\textbf{Output:} [(surface, Design, smooth, Positive), (null, Design, nice, Positive), (apps, Software, null, Negative)].''}

Finally, for each part, we start with simple content and iteratively refine it for better results. We find that the few-shot examples have a significant effect on the results. Therefore, we investigate different methods to choose few-shot examples, as presented in Section~\ref{subsec:fse}.

\subsection{Few-Shot Example Discovery for In-Context Learning}\label{subsec:fse}
In the few-shot setting, it is required to provide some examples for ChatGPT to do in-context learning. Because the prompt message for ChatGPT is limited in length, we cannot inject all examples in training data at a time. One question is how to select a few appropriate examples from training data for a given test example? The recent work~\cite{Min_Lyu_Holtzman_2022} shows that ``ground truth demonstrations are in fact not required—randomly replacing labels in the demonstrations barely hurts performance on a range of tasks.'' In the experiments, our observations are consistent. Therefore, this study does not consider the balance of labels, e.g., sentiment polarity. Instead, we utilize the KNN algorithm to identify few-shot examples. The KNN algorithm is a non-parametric method that works by finding a predetermined number of training samples closest in distance to a new point. For each test sample, the top-$k$ closest few-shot samples are retrieved and then used to fill in the prompt template for in-context learning.

Specifically, we explore two feature extraction methods to compute the cosine similarity between a given test example and examples from the training data. The two methods differ only in the vector representation. The first method involves extracting a term frequency inverse document frequency (TF-IDF) feature vector from an example. Without loss of generality, the feature vector of example $x$ is denoted as ${\bf v}$=TF-IDF($x$), in which each element is the product of term-frequency and inverse document frequency of a word in the vocabulary. The second method involves obtaining the embedding vector from pre-trained language models, e.g., BERT in our experiments, denoted as ${\bf v}$=BERT($x$), where each element represents a hidden feature. Further, the KNN-based selection method on few-shot examples is shown in Algorithm~\ref{alg:knn}.

\begin{algorithm}
	\caption{KNN-based selection method}
	\label{alg:knn}
	\begin{algorithmic}[1]
		\Require hyperparameter $k$, training example set $D=\{x_i\}$, and testing example $x$
		\Ensure top-$k$ closest examples
		\State ${\bf v}$=TF-IDF($x$) (or ${\bf v}$=BERT($x$))
		\For{each $x_i \in D$}
		\State ${\bf v}_i$=TF-IDF($x_i$) (or ${\bf v}_i$=BERT($x_i$))
		\State $s_i=\cos({\bf v}, {\bf v}_i)$
		\EndFor
		\State \Return $x_i$ with the largest $k$ similarities $s_i$.
	\end{algorithmic}
\end{algorithm}

\section{Experiments}\label{sec:exp}
We introduce the evaluation setup in Section~\ref{subsec:es} and result analysis in Section~\ref{subsec:ra}.

\subsection{Experimental Setup}\label{subsec:es}

\subsubsection{Datasets}
We collect four public available datasets to compare ChatGPT with specialized methods on the task of aspect-category-opinion-sentiment quadruple extraction. The work~\cite{Zhang_Deng_Li_2021} releases two datasets, namely \textit{Rest15} and \textit{Rest16}, that include implicit aspects but no implicit opinions and are constructed based on the SemEval Shared Challenges~\cite{Pontiki_Galanis_Papageorgiou_2015,Pontiki_Galanis_Papageorgiou_2016}. The study~\cite{Cai_Xia_Yu_2021} provide the other two datasets, namely \textit{Restaurant-ACOS} and \textit{Laptop-ACOS} with both implicit aspects and opinions. We adopt exactly the same splits on the four datasets for training, validation and testing as the references~\cite{Zhang_Deng_Li_2021,Cai_Xia_Yu_2021}.

\subsubsection{Compared methods}
We consider the methods evaluated in the references~\cite{Zhang_Deng_Li_2021,Cai_Xia_Yu_2021} as the baselines to compare them with ChatGPT.

\begin{table*}[htb]
	\centering
	\begin{tabular}{l|ccc|ccc}
		\hline
		\multirow{2}{*}{Methods}   & \multicolumn{3}{c|}{\textit{Rest15}}  & \multicolumn{3}{|c}{\textit{Rest16}} \\
		\cline{2-7}
		& Precision & Recall  & F1  & Precision & Recall  & F1 \\
		\hline
		HGCN-BERT-Linear           &   0.2443  & 0.2025  & 0.2215    &  0.2536   & 0.2403  & 0.2468 \\
		HGCN-BERT-TFM              &   0.2555  & 0.2201  & 0.2365    &  0.2740   & 0.2641  & 0.2690 \\
		TASO-BERT-Linear           &   0.4186  & 0.2650  & 0.3246    & 0.4973    & 0.4070  & 0.4477 \\
		TASO-BERT-CRF              &   0.4424  & 0.2866  & 0.3478    & 0.4865    & 0.3968  & 0.4371 \\
		GAS                        &   0.4531  & 0.4670  & 0.4598    & 0.5454    & 0.5762  & 0.5604 \\
		\textsc{Paraphrase}        &   \textbf{0.4616}  & \textbf{0.4772}  & \textbf{0.4693}    & \textbf{0.5663}    & \textbf{0.5930}  & \textbf{0.5793} \\
		\hline
		ChatGPT                    &   0.2966  & 0.3786  & 0.3326    & 0.3609    & 0.4693 & 0.4081 
		\\
		\hline	
	\end{tabular}
	\caption{The performance comparison on the two datasets with implicit aspects but without implicit opinions}
	\label{tab:f1}
\end{table*}

\begin{table*}
	\centering
	\begin{tabular}{l|ccc|ccc}
		\hline
		\multirow{2}{*}{Methods}   & \multicolumn{3}{c|}{\textit{Restaurant-ACOS}}  & \multicolumn{3}{|c}{\textit{Laptop-ACOS}} \\
		\cline{2-7}
		& Precision & Recall  & F1  & Precision & Recall  & F1 \\
		\hline
		DP-ACOS            & 0.3467 & 0.1508 & 0.2104 & 0.1304 & 0.0057 & 0.0800 \\
		JET-ACOS           & 0.5981 & 0.2894 & 0.3901 & 0.4452 & 0.1625 & 0.2381 \\
		TAS-ACOS           & 0.2629 & 0.4629 & 0.3353 & 0.4715 & 0.1922 & 0.2731 \\
		EC-ACOS  & \textbf{0.3854} & \textbf{0.5296} & \textbf{0.4461} & \textbf{0.4556} & \textbf{0.2948} & \textbf{0.3580} \\
		\hline
		ChatGPT            & 0.3839 & 0.4640 & 0.4202 & 0.2172 & 0.2765 & 0.2433 \\
		\hline	
	\end{tabular}
	\caption{The performance comparison on the two datasets with implicit aspects and opinions}
	\label{tab:f1-imp}
\end{table*}

\begin{itemize}
	\item HGCN-BERT-Linear: It is a pipeline model consisting of a hierarchical graph convolutional network (HGCN)~\cite{Cai_Tu_Zhou_2020} for jointly detecting the aspect category and sentiment polarity, followed by a BERT-based
	model~\cite{Devlin_Chang_Lee_2018} with a linear layer on top for extracting the corresponding aspect and opinion term. 
	
	\item HGCN-BERT-TFM: It replaces the linear layer of HGCN-BERT-Linear with a transformer block.
	
	\item TASO-BERT-Linear: It extends the target-aspect-sentiment model~\cite{Wan_Yang_Du_2020} with a linear
	layer on top to extract both aspect and opinion terms simultaneously and constructs a unified model to predict quadruples.
	
	\item TASO-BERT-CRF: It replaces the linear layer of TASO-BERT-Linear with a conditional random field layer.
	
	\item GAS: It adapts the generative aspect-based sentiment model~\cite{Zhang_Li_Deng_2021} to directly treat the sentiment
	quadruples sequence as the target for learning the generation model.
	
	\item \textsc{Paraphrase}: It is a novel modeling paradigm to cast the quadruple extraction task to a paraphrase generation process that jointly detects all four elements, i.e., the aspect, category, opinion and sentiment~\cite{Zhang_Deng_Li_2021}.
	
	\item DP-ACOS: It is one of the representative rule-based methods for aspect-opinion-sentiment triple extraction, and it has been adapted for the quadruple extraction task by first extracting all the aspect-opinion-sentiment triples, followed by assigning the aspect category for each extracted triple~\cite{Cai_Xia_Yu_2021}.
	
	\item JET-ACOS: It is an end-to-end framework for aspect-opinion-sentiment triple extraction~\cite{Xu_Li_Lu_2020}, and it has been adapted for the quadruple extraction task, similar to DP-ACOS.
	
	\item TAS-ACOS: It adapts the input transformation strategy of the target-aspect-sentiment model~\cite{Wan_Yang_Du_2020} to perform category-sentiment conditional aspect-opinion co-extraction, following by filtering out the invalid aspect-opinion pairs to form the final quadruples.
	
	\item EC-ACOS: It first performs aspect-opinion co-extraction, and then predicts the sentiment polarity of the extracted aspect-opinion pair candidates conditioned on each category~\cite{Cai_Xia_Yu_2021}.
\end{itemize}

\begin{figure}[!t]
	\centering
	\subfloat[\textit{Restaurant-ACOS}]{\label{subfig:restaurant-top-k}%
		\includegraphics[width=0.5\columnwidth]{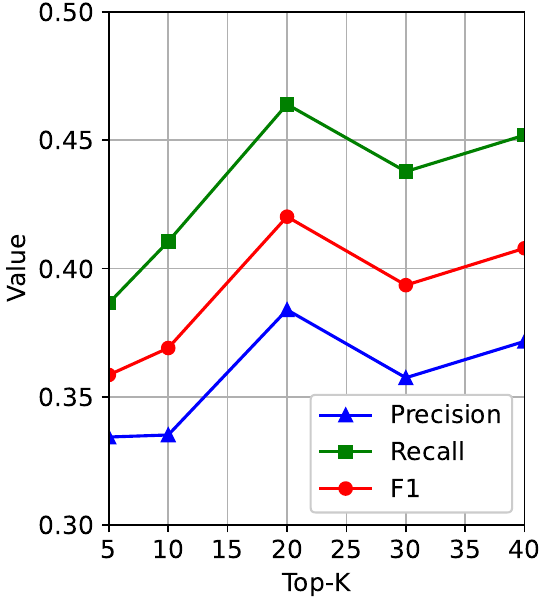}}
	\hfill
	\subfloat[\textit{Laptop-ACOS}]{\label{subfig:laptop-top-k}%
		\includegraphics[width=0.5\columnwidth]{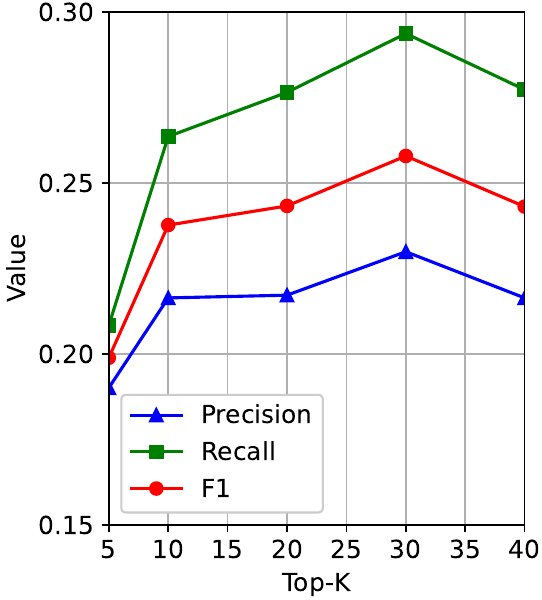}}
	\caption{Effect of few-shot examples}
	\label{fig:topk}
\end{figure}

\subsubsection{Evaluation metrics}
In line with the two references~\cite{Zhang_Deng_Li_2021,Cai_Xia_Yu_2021}, the Precision, Recall, and F1 scores are adopted as the main evaluation metrics. Moreover, we view a quadruple as correct if and only if the four elements as well as their combination are exactly the same as those in the ground-truth quadruples. Note that we report the results of compared methods from the original studies~\cite{Zhang_Deng_Li_2021,Cai_Xia_Yu_2021}.

\subsubsection{Settings}
To evaluate ChatGPT on the quadruple extraction task, we call API with a prompt message to the gpt-3.5-turbo, the most capable and cost-effective GPT-3.5 model\footnote{https://platform.openai.com/docs/models/gpt-3-5}. The temperature of API is set to 0 for making the outputs mostly deterministic and the other arguments are set by default. Unless otherwise specified, we apply the KNN algorithm with TF-IDF features to find the top-20 nearest examples from training data for a given test example in Section~\ref{subsec:fse}, and these few-shot examples are filled in the prompt template of \figurename~\ref{subfig:prompt-template}.

\subsection{Result Analysis} \label{subsec:ra}
We compare ChatGPT with baselines in Section~\ref{subsec:overcom}, investigate the effect of few-shot examples and selection methods in Sections~\ref{subsec:effect-example} and \ref{subsec:effect-select}, and conduct the relaxed study in Section~\ref{subsec:relaxed-study}.

\subsubsection{Overall comparison}\label{subsec:overcom}

\tablename~\ref{tab:f1} compares the performance of ChatGPT with those of the other methods reported in the work~\cite{Zhang_Deng_Li_2021} on the two datasets with implicit aspects but without implicit opinions. We have the following three findings: (1)~ChatGPT is much better than the pipeline models, including HGCN-BERT-Linear and HGCN-BERT-TFM, to some extent, which demonstrates the in-context learning ability of ChatGPT in quadruple extraction. (2)~According to the F1 score, ChatGPT is competitive with the TASO-based models, i.e., TASO-BERT-Linear and TASO-BERT-CRF; ChatGPT concentrates on recalling more quadruples, while the latter two focus on discovering more precise quadruples. (3)~ChatGPT is inferior to both GAS and \textsc{Paraphrase}, which are sequence-to-sequence T5 models, because ChatGPT just conducts in-context learning with few-shot examples but without updating model parameter, while the latter two apply the fine-tuning process on all the training data to optimize model parameters.

\tablename~\ref{tab:f1-imp} contrasts the performance of ChatGPT with the other methods reported in the work~\cite{Cai_Xia_Yu_2021} on the two datasets with implicit aspects and opinions. It is surprising that ChatGPT is competitive with the best method of EC-ACOS and much better than the other three baselines on the \textit{Restaurant-ACOS} dataset. The underlying reason is that ChatGPT is a generative model and has the intrinsic ability to deal with implicit aspects and opinions by representing them as a special token ``null''. Nonetheless, ChatGPT has a large gap in performance compared to EC-ACOS on the \textit{Laptop-ACOS} dataset because this dataset contains a much larger number of aspect categories than the former dataset, making it harder for ChatGPT to predict the correct one.

\begin{figure}[!t]
	\centering
	\subfloat[\textit{Restaurant-ACOS}]{\label{subfig:restaurant-select-method}%
		\includegraphics[width=0.5\columnwidth]{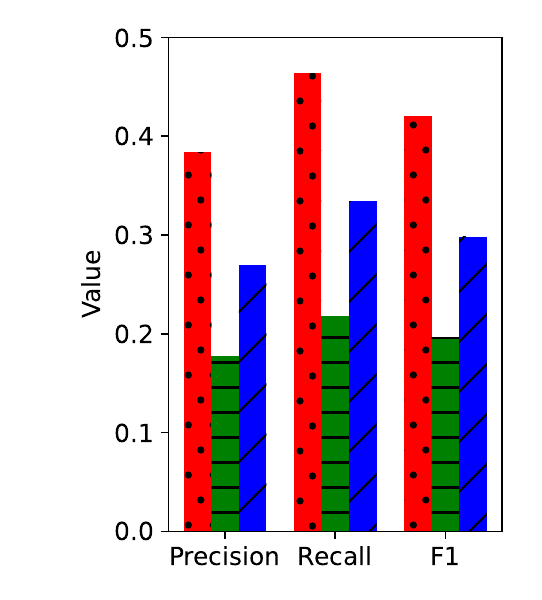}}
	\hfill
	\subfloat[\textit{Laptop-ACOS}]{\label{subfig:laptop-select-method}%
		\includegraphics[width=0.5\columnwidth]{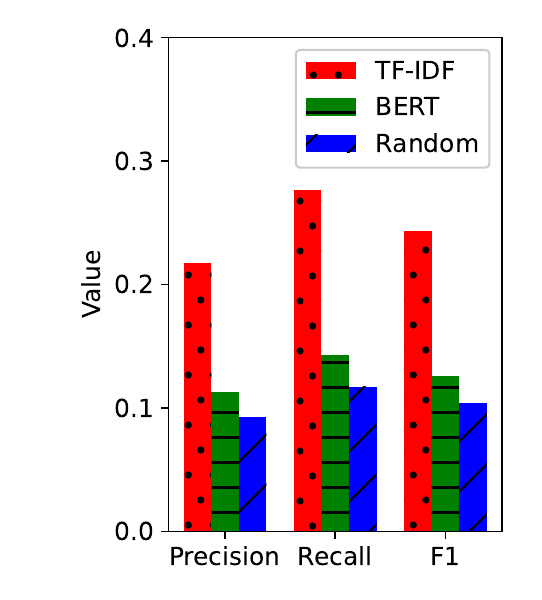}}
	\caption{Effect of selection methods on few-shot examples}
	\label{fig:sm}
\end{figure}

\subsubsection{Effect of few-shot examples} \label{subsec:effect-example}

Due to similar results, from now on, we will only report findings for the last two datasets with implicit aspects and opinions in the restaurant and laptop domains, respectively. \figurename~\ref{fig:topk} illustrates the effect of different numbers of few-shot examples on the performance of ChatGPT with in-context learning. As the number of few-shot examples increases from 5 to a certain value, i.e., 20 on the \textit{Restaurant-ACOS} dataset and 30 on the \textit{Laptop-ACOS} dataset, the performance gradually improves because these examples enable ChatGPT to conduct in-context learning; however, as the number of examples becomes larger, the performance unexpectedly fluctuates. Our explanation for this is that when the number of examples gets larger, the distance between the few-shot examples and the given test example becomes too far, which may introduce unrelated examples and result in unstable performance.

\subsubsection{Effect of selection methods} \label{subsec:effect-select}

\figurename~\ref{fig:sm} shows the effectiveness of ChatGPT with the effect of selection methods on few-shot examples. The selection method with TF-IDF achieves the best result because it can find more literally relevant examples than the other two selection methods. The random selection method is superior to the BERT-based selection method on the \textit{Restaurant-ACOS}, but the opposite is true on the \textit{Laptop-ACOS} dataset. The performance of a selection method is affected by two important factors: the relatedness of selected examples and inductive bias of the selection method. The \textit{Restaurant-ACOS} dataset has only 13 distinct categories. Even the random selection method can find related examples; moreover, it has weak inductive bias. Although the BERT-based selection method may find slightly more related examples, its strong inductive bias may be inconsistent with that of ChatGPT. As a result, in this case, the random selection method outperforms the BERT-based selection method. In contrast, the \textit{Laptop-ACOS} dataset has 121 nuanced categories, i.e., an order of magnitude larger than the \textit{Restaurant-ACOS} dataset. It is much harder for the random selection method to discover related examples, whereas the BERT-based selection method's ability to select more related examples compensates for its stronger inductive bias. Subsequently, in this case, the BERT-based selection method shows better performance than the random selection method.

\subsubsection{Relaxed study} \label{subsec:relaxed-study}

\begin{figure}[!t]
	\centering
	\subfloat[\textit{Restaurant-ACOS}]{\label{subfig:restaurant-iou}%
	\includegraphics[width=0.5\columnwidth]{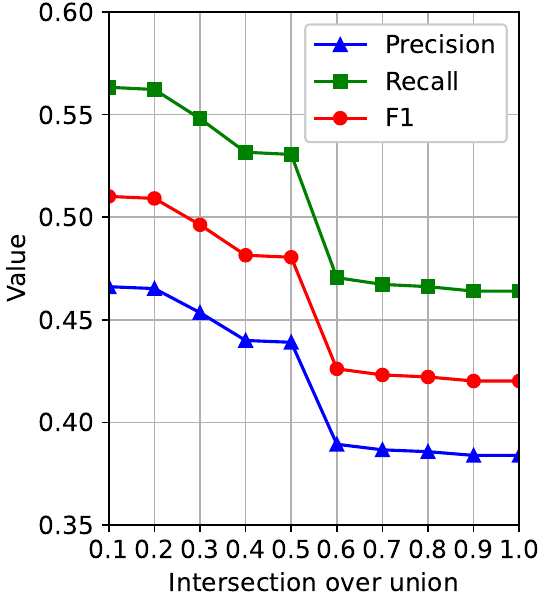}}
	\hfill
	\subfloat[\textit{Laptop-ACOS}]{\label{subfig:laptop-iou}%
	\includegraphics[width=0.5\columnwidth]{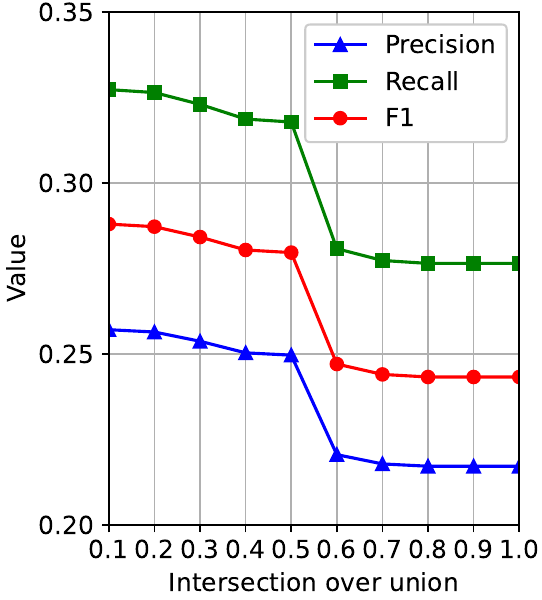}}
	\caption{Effect of intersection over union (IOU) of aspect and opinion terms}
	\label{fig:iou}
\end{figure}

From our experiments, we observe many cases, in which some function words cause inconsistency between the ground truth and the terms extracted by ChatGPT. For example, ChatGPT generates ``the surface'' and ``very dim'', but the ground truth gives ``surface'' and ``dim''. Thus, this motivates us to conduct a more relaxed study. We believe these inconsistent results can still indicate meaningful aspect/opinion terms. Here we calculate the intersection over union (IOU) of words between a predicted aspect term (or opinion term) and a ground-truth aspect term (or opinion term). We relax the condition of considering a predicted quadruple as correct from an IOU of 1 to a specified threshold, while maintaining the same other requirements. \figurename~\ref{fig:iou} depicts the effectiveness of ChatGPT with respect to the change of IOU thresholds. As the IOU threshold decreases, as expected, the three metrics improve. However, the improvement is much larger at the 0.5 point than at the other points on both datasets. The reason is that the aspect or opinion term often contains one or two words, and there are many cases with only one common word out of two words. When the IOU threshold is lowered to 0.1, ChatGPT performs significantly better. It is worth emphasizing that this relaxation is minor, because the corresponding category, sentiment, and their combination are exactly the same.

\section{Conclusion}\label{sec:con}
In this work, we develop a specialized prompt template tailored to aspect-category-opinion-sentiment quadruple extraction and empirically investigate the language understanding ability of ChatGPT on this complex quadruple extraction task. Based on extensive quantitative studies, we observe four important findings: (1)~ChatGPT is competitive with some pipeline solutions but inferior to sequence-to-sequence models with fine-tuning. (2)~Few-shot examples can help ChatGPT improve its performance based on in-context learning, whereas too much examples may degrade its effectiveness. (3)~Moreover, the selection method on few-shot examples has a significant effect on the performance of ChatGPT. (4)~Finally, ChatGPT can be enhanced by relaxing the requirements on the condition of considering a predicted quadruple as correct. These findings contribute to a better understanding on the capabilities and limitations of ChatGPT in this task of aspect-category-opinion-sentiment quadruple extraction. 

\section*{Limitations}
There are mainly two potential limitations. On the one hand, we manually design several prompt templates based on the prompt engineering guide from OpenAI, rather than extensively search a huge amount of prompt templates to find the best one; the performance of ChatGPT on this quadruple extraction task may be improved by devising a better prompt template. On the other hand, ChatGPT is only evaluated on the quadruple extraction task and its performance on other ABSA tasks is unknown in comparison to current state-of-the-art baselines. Our future work includes studying a dedicated method to automatically generate prompt templates for aspect-category-opinion-sentiment quadruple extraction and evaluating the ability of ChatGPT in various ABSA tasks.

%\begin{table}
%\centering
%\begin{tabular}{lc}
%\hline
%\textbf{Command} & \textbf{Output}\\
%\hline
%\verb|{\"a}| & {\"a} \\
%\verb|{\^e}| & {\^e} \\
%\verb|{\`i}| & {\`i} \\ 
%\verb|{\.I}| & {\.I} \\ 
%\verb|{\o}| & {\o} \\
%\verb|{\'u}| & {\'u}  \\ 
%\verb|{\aa}| & {\aa}  \\\hline
%\end{tabular}
%\begin{tabular}{lc}
%\hline
%\textbf{Command} & \textbf{Output}\\
%\hline
%\verb|{\c c}| & {\c c} \\ 
%\verb|{\u g}| & {\u g} \\ 
%\verb|{\l}| & {\l} \\ 
%\verb|{\~n}| & {\~n} \\ 
%\verb|{\H o}| & {\H o} \\ 
%\verb|{\v r}| & {\v r} \\ 
%\verb|{\ss}| & {\ss} \\
%\hline
%\end{tabular}
%\caption{Example commands for accented characters, to be used in, \emph{e.g.}, Bib\TeX{} entries.}
%\label{tab:accents}
%\end{table}

%\section*{Limitations}
%EMNLP 2023 requires all submissions to have a section titled ``Limitations'', for discussing the limitations of the paper as a complement to the discussion of strengths in the main text. This section should occur after the conclusion, but before the references. It will not count towards the page limit.  

% Entries for the entire Anthology, followed by custom entries
\bibliography{allbibliography}
\bibliographystyle{acl_natbib}

%\appendix
%\section{Example Appendix}
%\label{sec:appendix}

\end{document}